\documentclass[conference]{IEEEtran}
\IEEEoverridecommandlockouts
\usepackage{cite}
\usepackage{amsmath,amssymb,amsfonts}
\usepackage{algorithmic}
\usepackage{graphicx}
\usepackage{subcaption}
\usepackage{textcomp}
\usepackage{xcolor}
\usepackage{mathstuff}
\usepackage{tikz}
\usepackage{MnSymbol}
\usepackage{subcaption}
\usepackage{multirow}
\usepackage{booktabs}
\usetikzlibrary{shapes,positioning,decorations.pathmorphing,fit,shapes,calc, chains}
\usepackage{pifont}
\usepackage{url}
\usepackage{hyperref}
\usepackage{algorithm2e}
\usepackage{cases}

\usepackage{chngcntr}
\begin{document}

\title{ITGPT: Generative Pretraining on Irregular Timeseries \thanks{This work was partially supported by the Wallenberg AI, Autonomous Systems and Software Program (WASP) funded by the Knut and Alice Wallenberg Foundation. The computations were enabled by the Berzelius resource provided by the Knut and Alice Wallenberg Foundation at the National Supercomputer Centre. Correspondence: antoinehonor@gmail.com.}}

\author{\IEEEauthorblockN{Antoine Honoré$^\star$, Ming Xiao$^\star$}\IEEEauthorblockA{$^\star$\textit{Division of Information Science and Engineering, KTH Royal Institute of Technology}, Stockholm, Sweden}
}

\maketitle

\begin{abstract}
Timeseries regression models often struggle to leverage large volumes of labeled multimodal data, particularly when the data are irregularly sampled or contain missing values. This is common in domains like healthcare and predictive maintenance, where data are collected from unreliable sources, and labeling requires expert knowledge or costly equipments. Transformer-based large language models have proven effective on structured data such as text through self-supervised learning (SSL) and generative pretraining (GPT) frameworks. However, such models lack the flexibility to efficiently process irregularly sampled multimodal timeseries data. In this paper, we introduce ITGPT, an attention-based architecture designed for handling multimodal, irregularly sampled timeseries by allowing training with both SSL losses and GPT-like objectives. We evaluate its performance on a healthcare task with the TIHM  dataset, and a predictive maintenance task with the CompX dataset. Our results demonstrate that ITGPT achieves state-of-the-art performance without requiring resampling, feature fusion or explicit data imputation. Furthermore, when labels are scarce, ITGPT effectively leverages unlabeled data through SSL and GPT training, outperforming the purely supervised approach. This represents an important step towards efficiently using large and unstructured timeseries datasets for practical inference tasks. The code is available on GitHub: \url{https://github.com/antoinehonore/itgpt}.
\end{abstract}


\section{Introduction}
Multimodality arises when multiple sources of information are used to characterize a system \cite{zongSelfSupervisedMultimodalLearning2025}.
In many practical cases, sources might be de-synchronized, with different sampling frequencies and missing data.
For instance, in predictive maintenance (PdM) the quality of a mechanical piece must often be established from sensors with various sampling frequencies, e.g. noise and vibration sensors \cite{zontaPredictiveMaintenanceIndustry2020}.
Similarly in the medical field, a patient's health status must be established using information from fast or slow sources, for instance electrocardiograms (up 500 samples per second) combined with blood gas analysis (one sample per day) \cite{raviDeepLearningHealth2017a}. 
Additionally, in these contexts, labels about the health status of a patient or a mechanical piece might require expert knowledge and/or expensive sensors which limits their availability for supervised learning schemes. 
Current models handle irregularly sampled multimodal data with computationally expensive or overly simplistic resampling, imputation and feature fusion approaches. 
These approaches lack the ability to predict future input data and thus, only allow supervised learning frameworks.

Current approach for processing multimodal data often consider use cases related to image, text or speech processing.
Attention-based models such as Transformers are established as strong models that allow training with few labels, using self-supervised learning (SSL) or generative pretraining (GPT) frameworks.
These frameworks are key towards prediction models utilizing large amount of unlabeled data in PdM or healthcare contexts \cite{wangSelfSupervisedLearningRemote2022, krishnanSelfsupervisedLearningMedicine2022a}.

SSL is a machine learning paradigm that leverages unlabeled data to learn representations of input data by generating supervisory signals from the data itself \cite{guiSurveySelfSupervisedLearning2024}. 
Unlike traditional supervised learning, SSL does not require manual annotations, making it highly scalable and particularly valuable for domains where labeled data is scarce or expensive to obtain. 
A common approach involves predicting parts of the data from other parts, e.g. predicting the future of a timeseries from its past. 
SSL has been key for advancing domains such as computer vision and natural language processing, often serving as a pretraining step before fine-tuning on downstream tasks. 
The success of SSL reflects its ability to uncover structure and semantics in raw data without external supervision.
GPT is a strategy in which models are initially trained in an unsupervised manner on large datasets, in order to learn the distribution of the input data, typically using SSL objectives. 
This method became prominent with the introduction of models where the transformer architecture was pretrained to predict the next token in a sequence, capturing a wide range of linguistic patterns and world knowledge \cite{radfordLearningTransferableVisual2021}.
After pretraining, the model can be fine-tuned on specific supervised tasks with relatively little additional data, leveraging the rich representations learned during the generative phase.
This two-step process has set new benchmarks and forms the backbone of many state-of-the-art natural language processing systems such as ChatGPT, Mixtral and DeepSeek.

In the context of irregularly sampled multimodal data, there are currently no model to efficiently leverage unlabeled data for learning in SSL or GPT frameworks.

\subsection{Related works} 
The problem of handling irregularly sampled multimodal data was addressed with Kalman filters in the context of linear state space models \cite{ganComparisonTwoMeasurement2001a}.
These approaches provide efficient closed form inference models, but rely on strong knowledge of the linear processes and measurement systems. 
Recurrent neural networks with input space augmentation was shown promising for handling unimodal irregularly sampled signals \cite{cheRecurrentNeuralNetworks2018a}. 
This nonetheless might lead to a too large input space dimension.
Other RNNs with neural ODEs for modeling continuous time were also shown relevant, but computationally expensive. 
Other kernel-based Gaussian processes were also investigated, but scale quadratically with the number of training samples \cite{pmlr-v70-futoma17a}.

ITNet \cite{honoreITNetIrregularTimeseries2025} can be used to produce a multivariate data timeseries with arbitrary time sampling instants.
This output time series can then be used as an estimator for some quantity of interest in a supervised learning framework.
However, this often requires large quantities of accurately labeled data thus ignoring possibly large datasets of unlabeled data.
While this is straight to do on regularly sampled timelines, there is currently no standard method for multimodal and irregularly sampled timeseries data.

\subsection{Contributions}
Our contributions are as follows:
\begin{enumerate}
    \item We propose a framework called ITGPT which extends on ITNet to allow GPT-like objectives with irregularly sampled multimodal input data.
    \item We evaluate our model on two recent datasets for a healthcare and predictive maintenance tasks.
    \item We study our model when varying key hyperparameters: the model depth and the mixing layers.
    \item We show experimentally that our model benefits from SSL or GPT training frameworks when a few labeled samples are available for training.
\end{enumerate}

\section{Methods}
In this section we start by introducing the timeseries regression problem and the multimodal input data in \ref{sec:probstatement}.
For completeness we introduce the ITNet model in \ref{sec:itnet}.
The ITGPT architecture is finally described in \ref{sec:itgpt}.
We use the notation  $\forall N\in\N,\quad[N]=\{1,\dots,N\}$.

\subsection{Problem formulation}\label{sec:probstatement}
We introduce the problem of timeseries regression based on multimodal and irregular timeseries.
Let $M\in\N$ the number of observed modalities.
Suppose that we have access to $N\in\N$ independent realizations of the modalities of a system, in addition to a target timeseries.
For an observation $i\in[N]$, the data of modality $m\in[M]$ are denoted: \begin{equation}
    \tilde X_m^{(i)}=[\tilde{\mathbf{x}}_{m,1}^{(i)},\dots,\tilde{\mathbf{x}}_{m,L_m^{(i)}}^{(i)}]^\top\in\R^{L_m^{(i)}\times d_m},
\end{equation}
where $(L_m^{(i)},d_m)\in\N^2$ denote respectively the number of samples and the data dimension, and $\tilde{\mathbf{x}}_{m,t}^{(i)}\in\R^{d_m}$.
In addition, we assume that we have access to a sorted sequence of sampling timestamps for each modality and observation: \begin{equation}\label{eq:timeline}
    \boldsymbol\tau_m^{(i)}=[t_{m,1}^{(i)},\dots,t^{(i)}_{m,L_m^{(i)}}]^\top\in\R^{L_m^{(i)}}.
\end{equation}
Importantly, all timestamps across modalities have the same time reference, i.e. identical timestamps encountered in different modalities refer to the same time instant. 

Let \begin{equation}
X_m^{(i)}=[\tilde X_m^{(i)},\boldsymbol\tau_m^{(i)}]\in\R^{L_m^{(i)}\times (d_m+1)},
\end{equation}
denote the matrix composed of the concatenation of the data and the timestamps.
Let $\mathcal{X}^{(i)}=\{X_m^{(i)}\}_{m=1}^M$ denote the collection of data from all modalities for observation $i$.
Let $\tilde{Y}^{(i)}\in\R^{L_y^{(i)}\times d_c}$ denote the target sequence for observation $i$, with $(L_y^{(i)},d_c)\in\N^2$ respectively its length and dimension.
The target variable comes with its own timeline $\boldsymbol\tau_y^{(i)}\in\R^{L_y^{(i)}}$ (similar to Eq.~\eqref{eq:timeline}).
We write $Y^{(i)}=[\tilde{Y}^{(i)},\boldsymbol\tau_y^{(i)}]\in\R^{L_y^{(i)}\times(d_c+1)}$.
Finally, a dataset with $N$ observations is denoted $\mathcal{D}=\left\lbrace\left(\mathcal{X}^{(i)},Y^{(i)}\right)\right\rbrace_{i=1}^N$.

For each realization $i$, the aim is to perform regression of $\tilde{Y}^{(i)}$ at time instants in $\boldsymbol\tau_y^{(i)}$, using data $\mathcal{X}^{(i)}$. 
Importantly, the regression must be causal. 
That is, the target variable at time $t$ must be estimated only using data prior to $t$.

\begin{figure*}
\centering
\begin{subfigure}[t]{0.17\textwidth}
    \centering
    \includegraphics[width=\textwidth]{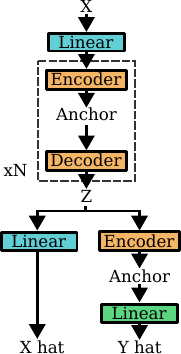}
     \caption{Multilayer ITGPT with prediction heads.}
    \label{fig:itgpt_overall}
\end{subfigure}
\begin{subfigure}[t]{0.67\textwidth}
    \centering
    \includegraphics[width=\textwidth]{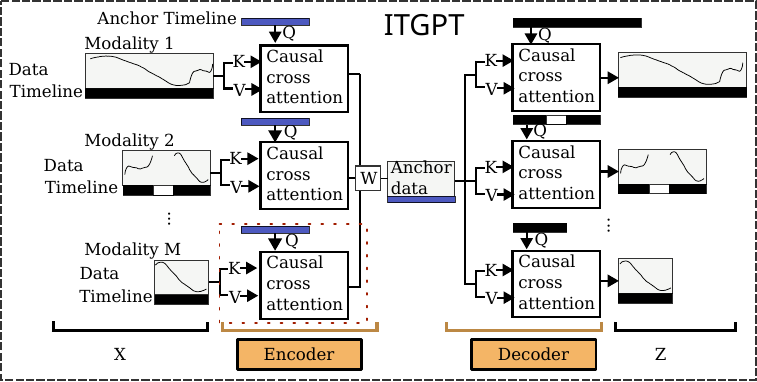}
    \caption{Single layer ITGPT. The encoder and decoder are based on ITNet. The causal cross attention and the computation of Q, K and V are further detailed in Fig.~\ref{fig:model_causal_inference}}
    \label{fig:itgpt_details}
\end{subfigure}
\caption{ITGPT architecture description. The encoder and decoders are two ITNet models. 
X and Z are multimodal data and the following linear layers are specific to each modality. 
Anchor denotes a timeseries of anchor data with a predefined timeline.}
\label{fig:itgpt}
\end{figure*}

\subsection{ITNet}\label{sec:itnet}
For completeness, this section introduces ITNet similarly to our previous work\cite{honoreITNetIrregularTimeseries2025}.
Let $t\in\boldsymbol\tau_y$ denote a time instant at which the target variable is sampled.
Let $\forall m\in[M],X_{m,t}$ denote the sub-matrix of rows of $X_m$ with the corresponding sampling times strictly prior to $t$;
similarly, $\boldsymbol\tau_{m,t}$ denote the sub-vector of $\boldsymbol\tau_{m}$ with time instants strictly prior to $t$.
The regression task consists in estimating the target sequence at all time $t\in\boldsymbol\tau_y$, using all data prior to $t$ across all modalities.
ITNet is formulated as follows with all available data up to time $t$:\begin{equation}\label{eq:estimator}
    \text{ITNet}\left(\{X_{m,t}\}_{m=1}^M\right)= W\left[g_1(t,X_{1,t}),\dots,g_M(t, X_{M,t})\right]\in\R^{d_c},
\end{equation}
where $\forall m\in[M],$ $g_m(.,.)$ are causal cross-attention mechanisms with output embedding dimension $d_o\in\N$, and a linear mixing layer $W\in\R^{d_c\times Md_o}$.
%The cross-attention mechanisms are based on functions to compute queries, keys and values, $f_Q$, $f_K$ and $f_V$ respectively (Details in Section\ref{sec:causal_mech}).

\begin{figure}[ht!]
    \centering
    \includegraphics[width=0.35\textwidth]{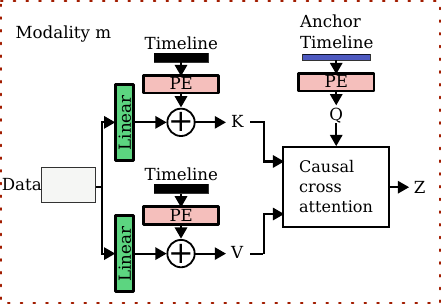}
    \caption{Causal cross-attention in ITNet. We use additive positional encoding for the computation of the key and value tensors from the data timeline. The positional encoding of the anchor timeline is directly used in the computation of the query tensor. PE: position encoding.}
    \label{fig:model_causal_inference}
\end{figure}
\subsubsection{Causal cross-attention}\label{sec:causal_mech}
In this section, we give a complete description of the modality-specific causal cross-attention mechanisms as proposed in \cite{vaswaniAttentionAllYou2017}, see also Fig.~\ref{fig:model_causal_inference}.
The output of the cross-attention for a timestamp $t$ is specified as a function of $t$ and of the modality data up to $t$, i.e. $g_m(t,X_{m,t})$.
In vector form, the $m$-th causal cross-attention mechanisms is expressed as follows: \begin{equation}
    g_m(t, X_{m,t}) = \sum_{t^\prime\in\boldsymbol\tau_{m,t}}\alpha_{t,t^\prime}^{(m)} \mbv_{t^\prime}\in\R^{d_o},
\end{equation}
where $\mbv_{t^\prime}\in\R^{d_o}$ denotes the row of $f_V(X_{m,t})$ corresponding to time instant $t^\prime$.

The coefficients $\alpha_{t,t^\prime}^{(m)}\in[0,1]$ are calculated with a similarity function, and so that no future information is used in the computation of the output at time $t$:\begin{equation}\label{eq:attentionweights}
\alpha_{t,t^\prime}^{(m)} = \left\lbrace\begin{matrix}\frac{\text{Sim}(\mbq_t,\mbk_{t^\prime})}{\sum_{s\in\boldsymbol\tau_{m,t}} \text{Sim}(\mbq_t,\mbk_{s})}&\text{ if } t^\prime<t,\\0&\text{ otherwise,}\end{matrix}\right.
\end{equation}
where the case $\mbk_{t^\prime}\in \R^{d_k}$ denotes the row of $f_K(X_{m,t})$ corresponding to time instant $t^\prime$. 
Similarly, let $\mbq_t\in \R^{d_k}$ denotes the row of $f_Q(\boldsymbol\tau_{y})$ corresponding to time instant $t$.
In vanilla attention, the similarity function is: \begin{equation}\label{eq:simfunction}
    \text{Sim}(\mbq_t,\mbk_{t^\prime}) = \exp\left(\frac{\mbq_t^T\mbk_{t^\prime}}{\sqrt{d_k}}\right) \in \R_+,
\end{equation}
where $d_k\in\mathbb{N}$ is the dimension of the keys and queries, i.e. the output dimension of $f_Q$ and $f_K$. 

\paragraph{Queries, keys and values}
To compute queries, $f_Q$ encodes $t$ in dimension $d_k$ (assumed an even number) using the position encoding (PE) function $p(t)$ as described in \cite{vaswaniAttentionAllYou2017}:
\begin{equation}
    \mbq_t=f_Q(t) = p(t)=\left[\dots, \sin\left(\omega_it\right), \cos\left(\omega_it\right),\dots\right]^T\in \R^{d_k},
\end{equation}
where $\forall i\in [\frac{d_k}{2}],~\omega_i=\Lambda^{-\frac{2i}{d_k}}$ and $\Lambda\in\R_+$ should be larger than the largest timestamp.
This position encoding scheme projects scalar timestamps to a cube: $[-1,1]^{d_k}$. 
This is done by sampling sines and cosines of different wavelengths. 
The scalar product following the position encoding in equation~\eqref{eq:simfunction} leads to timestamp differences projected in a similar way (See appendix).

The keys and values are computed from the data and with additive position encoding of the corresponding timestamps $t^\prime\in\boldsymbol\tau_m$:
\begin{equation}
    \mbk_{t^\prime}=f_K(\mbx_{m,t^\prime})= \tilde{\mbx}_{m,t^\prime}W^{K} + p(t^\prime)\in\R^{d_k},
\end{equation}
where $W^{K}\in\R^{d_m\times d_k}$ is trainable. 
Similarly:
\begin{equation}
    \mbv_{t^\prime}=f_V(\mbx_{m,t^\prime})= \tilde{\mbx}_{m,t^\prime} W^{V} + p(t^\prime)\in\R^{d_o},
\end{equation}
where $W^{V}\in\R^{d_m\times d_o}$ is trainable.

\subsection{ITGPT}\label{sec:itgpt}
ITNet cannot be used in a GPT framework because the output cannot directly be compared to its input.
In ITGPT, we chain together pairs of ITNets: \begin{itemize}
    \item a first ITNet (the encoder) is used with multimodal data as input and outputs a single multivariate timeseries with a predefined timeline (the "anchor" data).
    \item the second ITNet (the decoder) uses the anchor data as input and uses no output mixing layer, i.e. outputs multimodal data. 
\end{itemize}
The anchor data are similar to the target sequence of ITNet but serve a difference purpose: in ITGPT they are used to anchor, or synchronize all modalities on the same time sampling instants.
In the decoder, the initial data timelines are used for query computation (as opposed to the anchor timeline in the encoder), and the anchor data is used as input to the keys and values computation (as opposed to the input multimodal data).
This forms a model with multimodal data both in the input and in the output.
An arbitrary number $L$ of such encoding/decoding pairs can then be chained together to obtain a deep model with increased capacity.

The encoding/decoding operation at layer $l>0$ is formulated as follows:
\begin{numcases}{}
    A^{(l)} &$= \text{ITNet}\left(\{[E^{(l-1)}_m,\boldsymbol\tau_m]\}_{m=1}^M\right)\in\R^{L\times d_a}$,\label{eq:anchor_from_itnet}\\
    Z^{(l)} &$= Z^{(l-1)} + \varphi\left(A^{(l)}\right)\in\R^{L\times d_a}$,\label{eq:anchor_skip}\\
    E^{(l)}_m &$= g_m\left(\boldsymbol\tau_m, Z^{(l)}\right)\in\R^{L_m\times d_m},~\forall m\in[M]$,\label{eq:decoding_itnet}
\end{numcases}
where $\{E^{(0)}_m\}_{m=1}^M=\left\lbrace X_m\right\rbrace_{m=1}^M$, and a residual connection is introduced in \eqref{eq:anchor_skip} with $Z^{(0)}=0\in\R^{L\times d_a}$ and $\varphi$ a component-wise non-linearity, e.g. ReLU.
The model is depicted in Fig.~\ref{fig:itgpt}: with additional prediction and embedding transforms in Fig.~\ref{fig:itgpt_overall} and the details of an encoder/decoder pair in Fig.~\ref{fig:itgpt_details}.
Equation~\eqref{eq:decoding_itnet} can be viewed as an ITNet layer without output mixing layer. 

\section{Experiments}
We describe the datasets used to evaluate our ITGPT architecture in Section\ref{sec:datasets}.
Next, in \ref{sec:model_training} we present the training strategies and the loss functions used to test ITGPT abilities with a few labeled samples and SSL-like objectives. 
Section \ref{sec:hyperparameters} describes our architectures with the hyperparameters we fixed and the ones we varied.

\subsection{Datasets}\label{sec:datasets}

\subsubsection{Technology Integrated Health Management (TIHM)}
\paragraph{Description}
We evaluated our model using the TIHM dataset, an open dataset designed for remote healthcare monitoring in dementia \cite{palermoTIHMOpenDataset2023}. 
The dataset consists in longitudinal, in-home observation of $56$ people living with dementia (PLWD).
Data were collected by multiple connected sensors and medical devices.

\paragraph{Modalities}
In-home activity was monitored with passive infrared sensors, door sensors and under-mattress sleep mats, which provided minute level data on sleep states heart rate and respiration. 
Physiological were collected daily using commercially available devices measuring blood pressure, heart rate, body temperature, body weight, muscle mass and hydration.
We provide an overview of the data modalities and the different collection methods in Table~\ref{tab:modality_details_tihm}.

\paragraph{Labeling}
The dataset includes clinically validated health event labels. 
Alerts were generated through predefined thresholds and analytical models, and then verified by a monitoring team. 
Events include agitation episodes, abnormal blood pressure, body temperature, body water levels, heart rate, and weight changes. 
In addition, demographic information (age group, sex, dementia diagnosis) is provided for all participants but were not used in our model since the authors reported that they do not impact performances.

\paragraph{Performance metrics}
We report the binary sensitivity (Recall) and the specificity scores, similarly to the original paper. 
In addition, we report the AUROC to get a unique comparison metric based on specificity and sensitivity.
The metrics are reported in a timeseries split cross-validation. Roughly speaking, this scheme consists in training on the beginning of all patient timeseries and validating on the end (See \cite{palermoTIHMOpenDataset2023} for details).

\subsubsection{SCANIA Component X}
We used the CompX dataset, a publicly available real-world multivariate time series dataset designed for predictive maintenance research in heavy-duty trucks \cite{kharazianSCANIAComponentDataset2025a}. 

\paragraph{Description}
The dataset comprises sensor-derived operational data, maintenance records, and truck specification information collected from a fleet of 28596 SCANIA vehicles. 
The underlying system "CompX", remains anonymized to protect proprietary information.
\begin{itemize}
    \item The operational data, recorded via onboard sensors and extracted from vehicle control units, are formatted as a multivariate time series, capturing temporal dynamics through unevenly sampled readouts.
    These include both numerical counters and histogram-encoded variables, the latter offering compact summaries of feature distributions across bins defined by domain-informed thresholds. The different variables are depicted in Fig~\ref{fig:example_data_compx}.
    \item Vehicle specifications, such as engine type and configuration, are included as categorical variables and free from missing values.
\end{itemize}
Time progression is represented by usage intervals for CompX.

\paragraph{Modalities}
The training set contains $107$ anonymized features, available at irregular time steps for each vehicle. 
The features include: six histogram-based variables (with up to 36 bins); eight univariate numerical counters; categorical vehicle specifications encoded as one-hot vectors leading to binary vectors of dimension $94$.
All these variables were used as individual modalities, which leads to a total of $M=15$ modalities with varying sampling timelines and dimensions, a summary is available in Table~\ref{tab:modalities_details}.

\begin{figure}[ht!]
    \centering
    \includegraphics[width=1\linewidth]{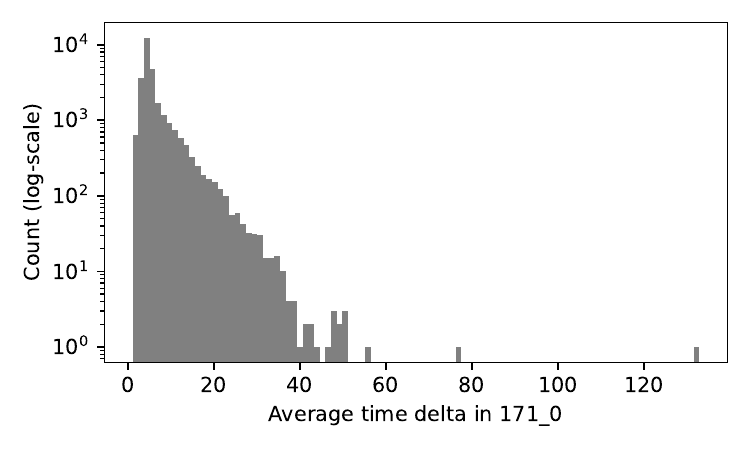}
    \caption{Histogram of average time deltas between samples of modality "171\_0" across all vehicles in the training dataset.}
    \label{fig:histogram_average_delta_171_0}
\end{figure}

A challenge in the dataset is the irregularity in time deltas between consecutive time points. 
We illustrate this on one modality in Fig~\ref{fig:histogram_average_delta_171_0}.

\paragraph{Labeling}\label{sec:labeling}
Maintenance records were compiled from official SCANIA workshop logs, including repair invoices and service orders. 
These records identify whether CompX was repaired or replaced during the observation period for each vehicle. 
For modeling purposes, the component is considered failed at time $t$ if any repair or replacement of CompX was documented at time $t$; otherwise, it is considered healthy.
Each time step in a vehicle recording is labeled with one of six classes, Class 1: No failure within 48h; Class 2: Failure within 24h to 48h; Class 3: Failure within 12h to 24h; Class 4: Failure within 6h to 12h; Class 5: Failure within 6h; Class 6: Potential failure within 48h (censored data).

\paragraph{Performance metrics}
For the CompX dataset, our performance metric for the validation sets is the area under the precision recall characteristic (AUPRC).
This metric is the most adequate in imbalanced multiclass classification problems, over, for instance, raw accuracy or AUROC, which tend to neglect disproportionate amounts of false positives over true positives. 
Additionally, the metric is threshold independent, which makes its calculation independent from potential poor calibrations across several folds.
AUPRC is extended to multiclass by averaging one-vs-rest binary AUPRC scores corresponding to each class.

The metrics are computed in a $5$-fold cross-validation framework to evaluate the ability of the models to generalize.

\subsection{Model training} \label{sec:model_training}
Our models are trained with backpropagation using the ADAM optimizer.
We use different training schemes to study the performances of ITGPT in settings where very few labels are available. 
First, we evaluate a pure SSL scheme, denoted ``CE+SSL'', where all the training dataset is used to compute a SSL loss, and only a few labels are used to compute a supervised learning loss ``CE''.
Second, we evaluate a GPT scheme, denoted ``GPT$\rightarrow$CE'', where the model is first trained only on the GPT loss, and then is finetuned on the supervised learning loss.

\subsubsection{Supervised learning loss}
Our supervised training loss is the standard cross-entropy loss between true and estimated labels:
\begin{equation}
    l_{\text{CE}}(\mby,\hat{\mby}) = \sum_k \mby_k\ln \hat{\mby}_k
\end{equation}

\subsubsection{SSL and GPT loss}
ITGPT allows to use a self-supervised learning loss in input data space, in this paper we experiment with the MSE loss of a one-step ahead prediction of the input data:
\begin{equation}
    l_{\text{MSE}}\left(\mathcal{X},\hat{\mathcal{X}}\right) = \sum_{m=1}^M \frac{1}{L_m}\sum_{t\in\boldsymbol\tau_m} ||\tilde{\mbx}_{t,m} - \hat{\tilde{\mbx}}_{t,m}||_2^2,
\end{equation}
where $\forall m \in \times [M],\forall t\in\{2,\dots,T_m\}$, $\hat{\tilde{\mbx}}_{t,m}$ is the output of ITGPT computed from data strictly prior to $t$. 

For the training scheme denoted "CE+SSL", the loss for a training sample $i$ is the sum of the MSE loss on in input data space and the CE loss only if the label is made available : \begin{equation}\begin{aligned}
    l_{\text{CE+SSL}}(\mathcal{X}^{(i)},\hat{\mathcal{X}}^{(i)}) =&\\
    l_{\text{MSE}}(\mathcal{X}^{(i)},\hat{\mathcal{X}}^{(i)}) &+ 1(i\in\mathcal{L}).l_\text{CE}(\mby^{(i)},\hat{\mby}^{(i)}),
    \end{aligned}\end{equation}
where $\mathcal{L}$ is the set with the indices of training samples with available labels, and $1$ is the indicator function.

For the training scheme denoted "GPT$\rightarrow$CE", the model is first pretrained with the MSE loss on $2$ epochs, and then trained only on the CE loss with the data with the available labels for $5$ epochs, i.e. for epoch $n\in[7]$, the loss function is formulated: \begin{equation}
    l_{\text{GPT$\rightarrow$CE}}(\mathcal{X}^{(i)},\hat{\mathcal{X}}^{(i)})=\begin{cases}
       l_\text{MSE}(\mathcal{X}^{(i)},\hat{\mathcal{X}}^{(i)}) & \text{if } n\leq 2,\\
        l_\text{CE}(\mby,\hat{\mby})&\text{if } n>2, i\in\mathcal{L},\\
        0&\text{if } n>2, i\notin\mathcal{L}.
    \end{cases}
\end{equation}

\subsubsection{Censored data in CompX}
Class 6 in the original CompX dataset accounts for censored data (As described in \ref{sec:labeling}).
This class is ambiguous in the sense that samples from classes $1$ to $5$ might be labeled as $6$. 
Training with a CE loss for the samples in this class will thus make the dataset harder to fit, since all samples could be classified as $6$. 
This is illustrated in Fig.~\ref{fig:example_pred_cm} where many more samples from class $6$ are misclassified into the other classes equally 
We thus ignore samples with class $6$ when computing a $CE$ loss. 
The input data are still used without labels in the SSL and GPT training schemes where the MSE loss does not dependently on the label.

\subsection{Hyperparameters}\label{sec:hyperparameters}
A list of the hyper-parameters is available in Table~\ref{tab:hyperparameters}.
The results of various architectural choices are reported on the CompX dataset because it is the dataset with the most complexity in terms of number of samples and .

\subsubsection{Model architecture}
We study the impact of varying the architecture of two important components of the model: the depth and the mixing layers.
The mixing layer corresponds to the layer which outputs the anchor data from the encoding of the data from individual modalities (See Fig. \ref{fig:itgpt_details}). 
We experiment with linear mixing layers, as well as multilayer perceptron layers with $1$ and $2$ hidden layers.
The depth corresponds to the number of chains of pairs of ITNet models, with a depth of $1$ corresponding to an ITNet encoder alone. 
We experiment with depth from $1$ to $7$ pairs of encoder/decoder, and the models of increasing depth are expected to overfit data.
We also add experiments varying the percentage of dropout during training, in an attempt to control the overfitting.
To reduce the number of possible combinations of hyper-parameters, we found reasonable values for other parameters. 

\begin{table}[ht!]
    \centering
    \begin{tabular}{c|c|p{0.8cm}|c}
        &Short Description & Symbol & Value\\\hline
       Fixed &key/queries dimensions  & $d_k$ & $32$ \\
       &key/query/value maps & $f_{Q/K/V}$ & Linear\\
        &Activation function  & $\varphi$& ReLU\\
        &Anchor dimension & $d_a$ & $64$\\
        &Batch size & $B$ & $64$\\
        &Optimizer& /& ADAM\\
        &Learning rate & $\lambda$ & $5\times 10^{-4}$\\
        &Number of epochs& $N_e$& 20\\\hline
      Varied  & Chain depth & $L$ & $\{1,2,3,4,5,6,7\}$\\
        &Dropout probability & $p$ & $\{0,0.1,0.2,0.3\}$\\
        & Mixing layer & $W$ & \{Linear, MLP/1, MLP/2\}\\ 
        & \% used labels & $p_l$ & $\{0.1\%,0.2\%,\dots,1\%\}$
    \end{tabular}
    \caption{List of fixed and varied hyperparameters for the CompX dataset.}
    \label{tab:hyperparameters}
\end{table}

\begin{figure*}[ht!]
    \begin{subfigure}[t]{0.49\textwidth}
        \centering
        \includegraphics[width=\textwidth]{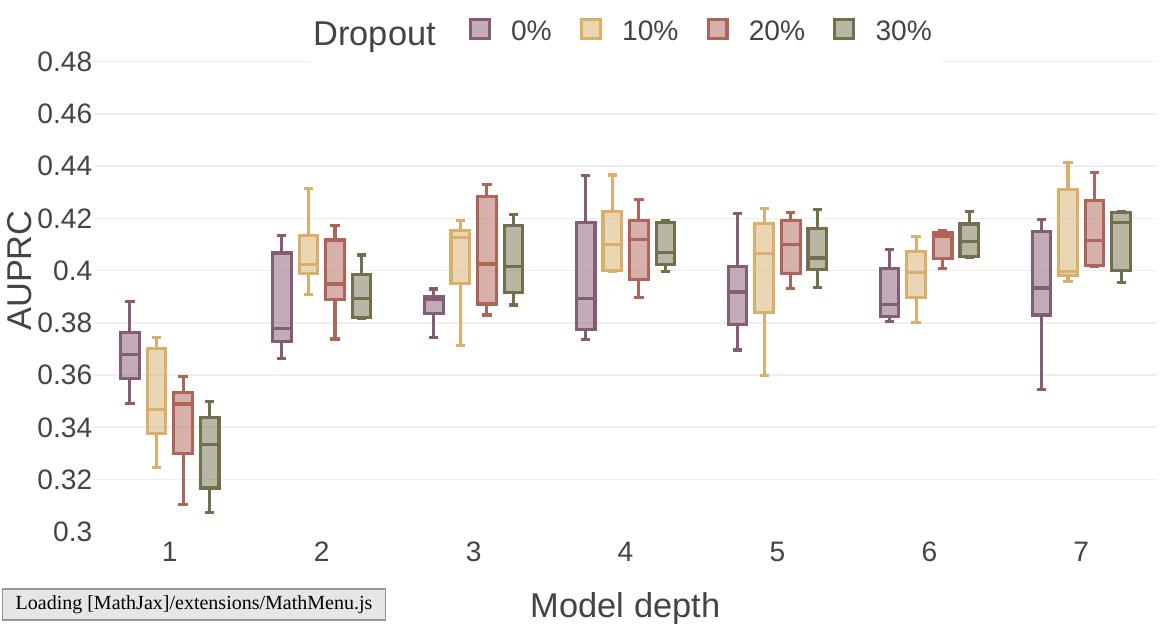}
        \caption{Comparing levels of dropout for various model depth with linear mixing layers.}
        \label{fig:vs_n_layers_linear}
    \end{subfigure}\hfill
    \begin{subfigure}[t]{0.49\textwidth}
        \centering
        \includegraphics[width=\textwidth]{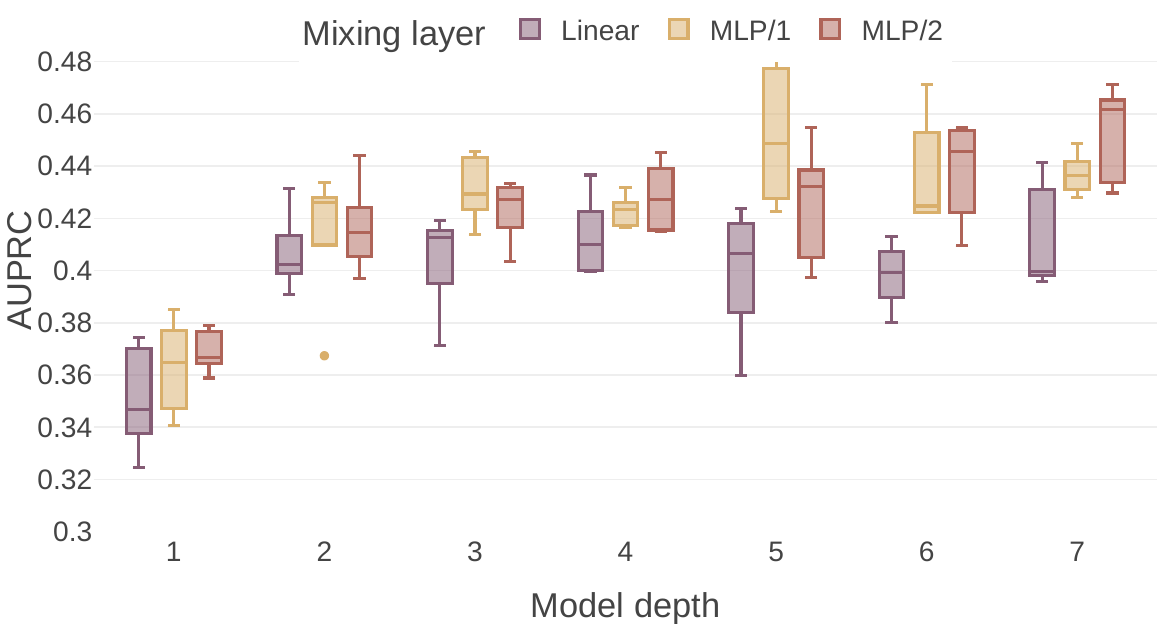}
        \caption{Comparing linear, MLP/1 and MLP/2 mixing layers for various model depth, with dropout=10\%.}
        \label{fig:vs_n_layers_dropout10}
    \end{subfigure}\\
    \begin{subfigure}[t]{0.49\textwidth}
        \centering
        \includegraphics[width=\textwidth]{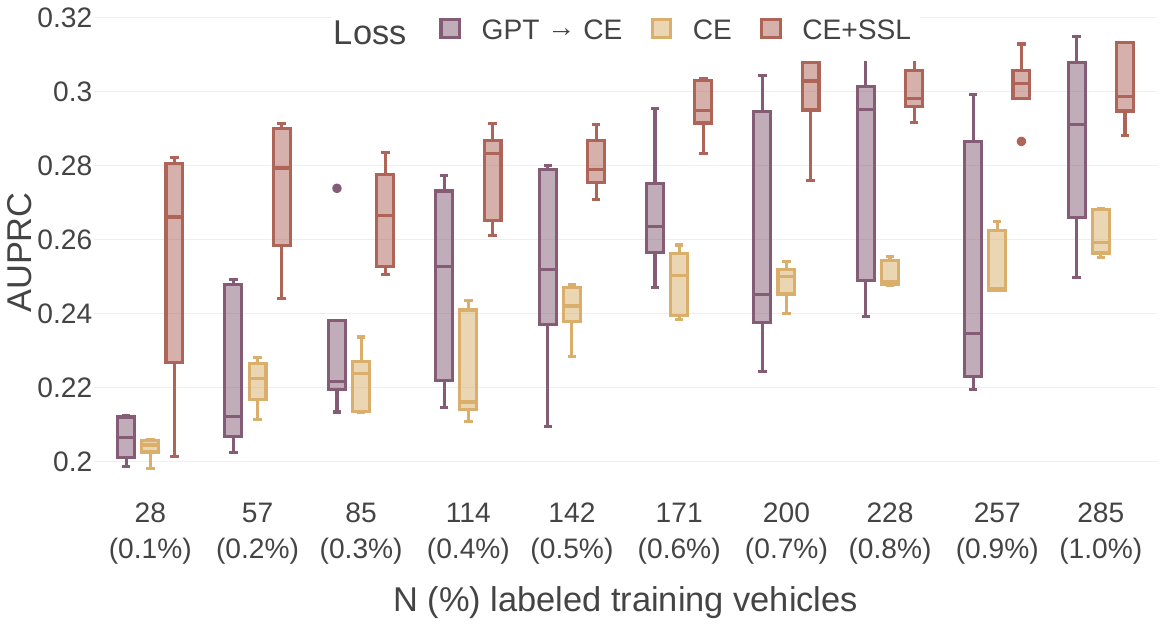}
        \caption{Performances in a small labeled dataset setting. 
        CE: Training only with cross-entropy; 
        CE+SSL: Training with both cross entropy and a self-supervised loss.;
        GPT$\rightarrow$CE: Pretraining with the self-supervised loss, fine tuning with cross-entropy loss.}
        \label{fig:boxplot_CESSL}
    \end{subfigure}\hfill
    \begin{subfigure}[t]{0.49\textwidth}
        \centering
        \includegraphics[width=0.7\linewidth]{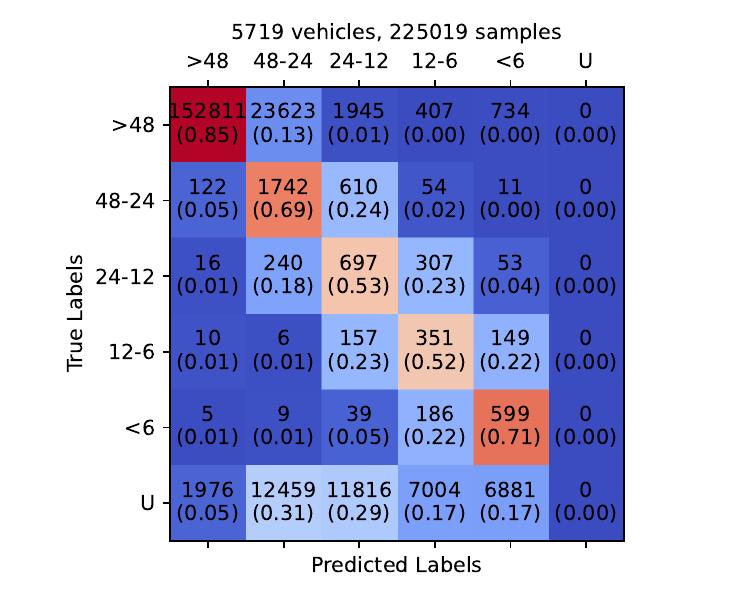}
        \caption{Confusion matrix on a validation fold for a model trained to predict $5$ classes. Class 6 attracts samples from all other classes and thus, should not be used for training.}
        \label{fig:example_pred_cm}
    \end{subfigure}
    \caption{Results in a $5$-fold cross-validation framework on the CompX dataset.}
    \label{fig:compx_results}
\end{figure*}

\section{Results}
We start with the results on the TIHM dataset in Section~\ref{sec:results_tihm}. 
The results of experiments with various architecture choices on the CompX dataset are presented in Section~\ref{sec:res_varying_architectures}. 
We then show and discuss the results of our experiments with fewer labeled samples in Section~\ref{sec:res_amount_labels}. 
The results are reported in box plots depicting the median and first and third quartiles. 
The experiments are seeded so that the folds are identical across hyperparameter choices. 

\subsection{TIHM dataset}\label{sec:results_tihm}
We start with the results obtained on TIHM with several losses and all the available labeled data. 
The results are shown in Table~\ref{tab:tihm_lossescomparison} for several losses. 
We also mention that the original paper \cite{palermoTIHMOpenDataset2023} reports specificity and sensitivity of around $0.8$ with a linear model and linear interpolation.
Table~\ref{tab:tihm_lossescomparison} shows that our model performs similarly, although slightly under the performance reported by the original authors. 
We mention this is done entirely from the raw data and without the need for interpolation. 

The GPT+CE training method does not improve performances over the classical CE loss function.
However, we see that the CE+SSL framework improve the mean recall performance marginally from $0.63$ to $0.73$.
The other metrics, specificity, AUROC, and F1score remain similar.

\begin{table}[ht!]
    \centering
    \begin{tabular}{lllll}
\toprule
 & Recall & Spec & AUROC & F1score \\
Loss &  &  &  &  \\
\midrule
CE & 0.68 (0.14) & 0.77 (0.08) & 0.78 (0.05) & 0.25 (0.11) \\
CE+SSL & 0.73 (0.17) & 0.76 (0.09) & 0.79 (0.06) & 0.26 (0.11) \\
GPT → CE & 0.6 (0.22) & 0.78 (0.1) & 0.76 (0.06) & 0.23 (0.12) \\
\bottomrule
\end{tabular}

    \caption{Numerical results for several binary classification metric. }
    \label{tab:tihm_lossescomparison}
\end{table}

Next we discuss the results when the training scheme and the number of available samples vary.
Overall, as expected, the recall performance decrease and the specificity increases as we train with an increasing number of labels.
We see only marginal differences between the training schemes, particularly for the specificity.
Interestingly, the recall performance of the ``CE+SSL" scheme remains stable after $25$ labeled samples, while both ``CE" and ``GPT$\rightarrow$CE" performances keep decreasing.
This is coherent with the mean performances in the table reported previously.

\begin{figure}[ht!]
    \begin{subfigure}{0.9\columnwidth}
    \centering
        \includegraphics[width=\linewidth]{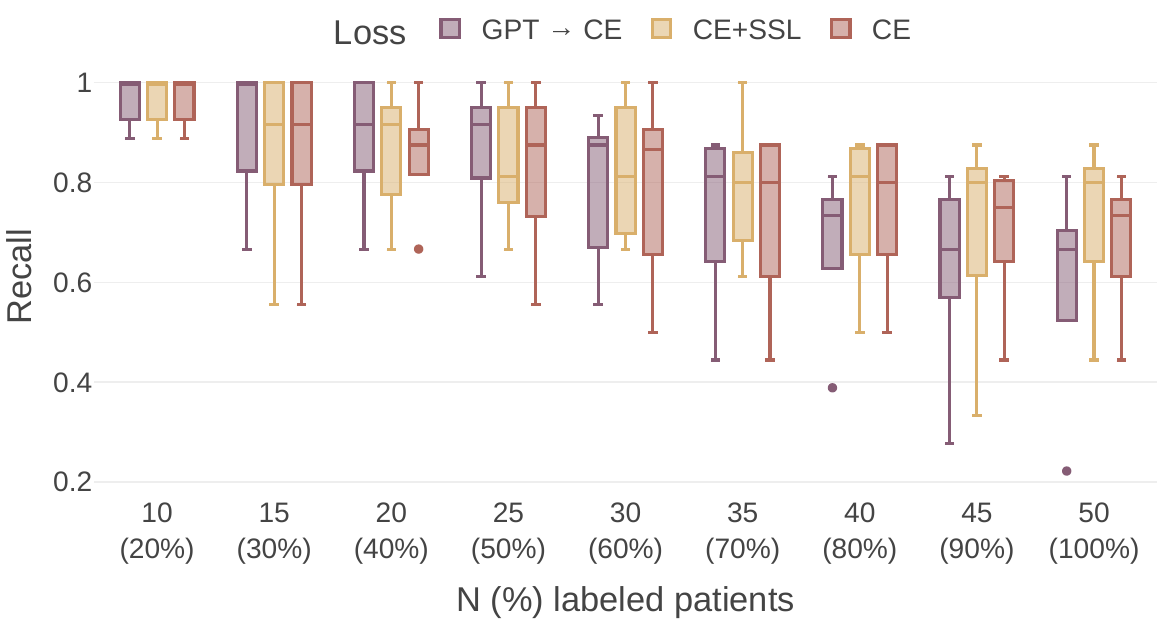}
        \caption{Recall/Sensitivity performance}
        \label{fig:results_tihm_recall}
    \end{subfigure}\\
    \begin{subfigure}{0.9\columnwidth}
        \centering
        \includegraphics[width=\linewidth]{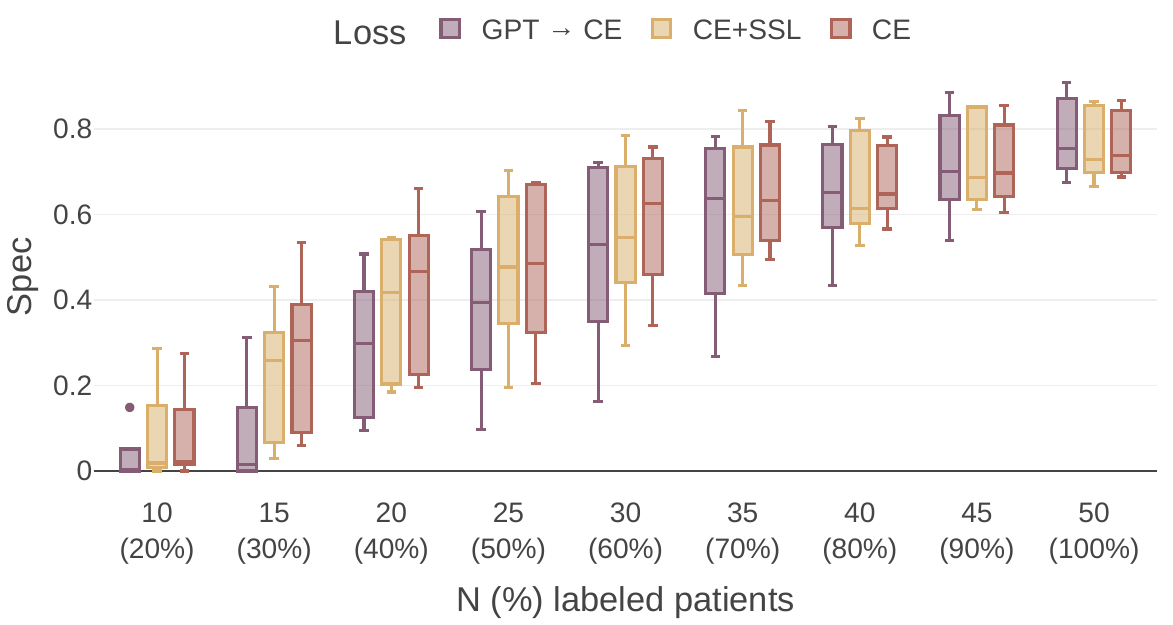}
        \caption{Specificity performance}
        \label{fig:results_tihm_spec}
    \end{subfigure}
    \caption{Recall (Sensitivity) and Specificity  for an increasing amount of available labels.}
    \label{fig:placeholder}
\end{figure}

\subsection{SCANIA component X (CompX) dataset}
\subsubsection{Varying ITGPT architectures}\label{sec:res_varying_architectures} 
We start by presenting the results obtained on the full CompX dataset with a CE loss while varying the dropout probability, the chain depth and the type of mixing layer.

Fig.~\ref{fig:vs_n_layers_linear} shows that increasing the probability of dropout has a detrimental effect at depth 1 and a beneficial effect at depth larger than 2. 
This shows that overfitting must be controlled at large depth, in turn confirming that our architecture does not suffer from gradient collapse. 
This is important to be able to scale the capacity of the model for complex datasets.
A similar conclusion can be drawn for MLP/1 and MLP/2 mixing layers on Fig.~\ref{fig:vs_n_layers_mlp1} and Fig.~\ref{fig:vs_n_layers_mlp2}. 
An additional observation for these mixing layers is that dropout probability $>30\%$ seem to decrease performances compared to $10\%$ and $20\%$. 
This is a different behavior from the linear mixing layer case. 

In our setup, several choices of architectures achieve similar performance around $0.44$ AURPC score:
MLP/2 mixing layers, dropout probability $10\%$ and depth $3$ or $6$; as well as the MLP/1 mixing layers, dropout probability $20\%$ and depth of $6$ or $7$.
The results for linear mixing layers are slightly lower with the best-performing model at $0.42$ AUPRC with a depth of $4$ and a dropout probability of $20\%$.

Overall, this experiment confirms the sanity of our ITGPT architecture and stresses the importance of controlling overfitting when the model capacity increases.

\subsubsection{Varying the amount of available labels}\label{sec:res_amount_labels}
Both our GPT$\rightarrow$CE and CE+SSL training schemes improve over the simple CE loss scheme (Fig.~\ref{fig:boxplot_CESSL}).
We note that the performances of "CE+SSL" increase faster with the amount of available data compared to "GPT$\rightarrow$CE", with the highest score achieved when $85$ samples are available.
The lowest performance our models obtain is around $0.2$ AUPRC when only $28$ samples are labeled. 
When the amount of labels increases, the performances of the models trained with the CE loss only increases as expected.
The performances of the models trained with both GPT$\rightarrow$CE and CE+SSL also increases but only until $171$ available labels. 
With more available samples, the performances of GPT$\rightarrow$CE decreases slightly, and the performances of CE+SSL saturates a little under $0.28$ AUPRC.

\section{Conclusion}
In this work, we introduced ITGPT, a transformer-based architecture specifically designed to handle multimodal, irregularly sampled timeseries data without requiring resampling, feature fusion, or imputation. ITGPT leverages both self-supervised learning (SSL) and generative pretraining (GPT-style) objectives, making it especially effective in low-label regimes.
Our experiments demonstrate that ITGPT achieves state-of-the-art performance on a predictive maintenance task using the CompX dataset, particularly when labels are scarce. We show that SSL and GPT pretraining significantly boost performance over purely supervised learning, with the CE+SSL scheme benefiting the most from increasing amounts of labeled data.
We also explored architectural variants, revealing that dropout plays a critical role in controlling overfitting at deeper depths, and that MLP-based mixing layers outperform linear ones under optimal settings. Our best configurations reach AUPRC scores of 0.44, validating the capacity and robustness of the ITGPT architecture.
Finally, we identify several promising directions for future research, including:\begin{itemize}
    \item Improving interpretability through unitary mixing layers that track individual modality contributions;
    \item Reducing computational complexity, especially via more efficient attention mechanisms and modality-parallel CUDA kernels;
    \item Exploring sparse dot-product computations to optimize batching and reduce unnecessary inter-sample attention.
\end{itemize}
Overall, ITGPT offers a strong and flexible foundation for tackling real-world timeseries prediction tasks, and opens possibilities to leverage large and unstructured datasets in industrial or health care applications.

\section{Acknowledgements}
The authors wish to thank Baptiste Cavarec for valuable discussions on experimental design and insightful feedback during results interpretation.
We used ChatGPT for suggesting rephrasings of the abstract and conclusion.

\pagebreak
\newpage
\bibliography{itgpt}
\bibliographystyle{ieeetr}

\appendix
\renewcommand\thefigure{\thesection.\arabic{figure}}   
\setcounter{figure}{0}
\renewcommand\thetable{\thesection.\arabic{table}}   
\setcounter{table}{0}

\subsection{Position encoding}
The samples can be irregularly spaced across observations or modalities.
Let \begin{equation}
    \boldsymbol\delta_{m}^{(i)}=[0,t_{m,2}^{(i)}-t_{m,1}^{(i)},\dots,t_{m,L_m^{(i)}}^{(i)}-t_{m,L_m^{(i)}-1}^{(i)}]^\top\in\R_+^{L_m^{(i)}},
\end{equation} denote the vector of time difference between two consecutive timestamps.

\subsection{Additional CompX Data Information}

\subsubsection{Normalization}
Perturbation strategies were used to ensure data privacy and commercial confidentiality. The original timestamps and the true variable names were removed from the data, and feature values were scaled with an undisclosed factor.
According to the authors, the dataset retains its utility for a wide range of machine learning tasks, including classification, regression, anomaly detection, and survival analysis.
The perturbation applied to the dataset leads to features in the range $[0, 10^9]$ across all vehicles.
Scaling with large factors might lead to underflow since the range varies quite a lot per vehicle. 
Thus, we chose to apply a logarithmic normalization: $x \mapsto \ln\left(1+x\right)$.

\subsubsection{Labeling}
To ensure label quality, only vehicles with a complete service history within the SCANIA workshop network were included. 
This restriction avoids ambiguities arising from missing third-party repair data and helps reduce label noise. 
However, it does introduce selection bias toward vehicles that consistently use SCANIA-authorized maintenance services.

\begin{table}[ht!]
    \centering
    \begin{tabular}{p{0.8cm}p{1cm}p{1.5cm}p{2.1cm}}
\toprule
Modality name & Dimension & Total number of samples & \% vehicles with complete modality \\
\midrule
specs & 94 & 28.60K & 1.00 \\
397 & 36 & 1.32M & 0.80 \\
459 & 20 & 1.31M & 0.68 \\
291 & 11 & 1.31M & 0.68 \\
158 & 10 & 1.32M & 0.80 \\
272 & 10 & 1.32M & 0.81 \\
167 & 10 & 1.32M & 0.81 \\
370\_0 & 1 & 1.31M & 0.74 \\
835\_0 & 1 & 1.32M & 0.82 \\
309\_0 & 1 & 1.32M & 0.82 \\
837\_0 & 1 & 1.32M & 0.82 \\
427\_0 & 1 & 1.31M & 0.74 \\
666\_0 & 1 & 1.32M & 0.82 \\
171\_0 & 1 & 1.32M & 0.82 \\
100\_0 & 1 & 1.31M & 0.74 \\
\bottomrule
\end{tabular}

    \caption{Description of the modalities in the training dataset of $28596$ vehicles. Modalities are considered complete for a vehicle if the sampling and label timelines coincide.}
    \label{tab:modalities_details}
\end{table}

\begin{figure}[ht!]
    \centering
    \includegraphics[width=0.5\textwidth]{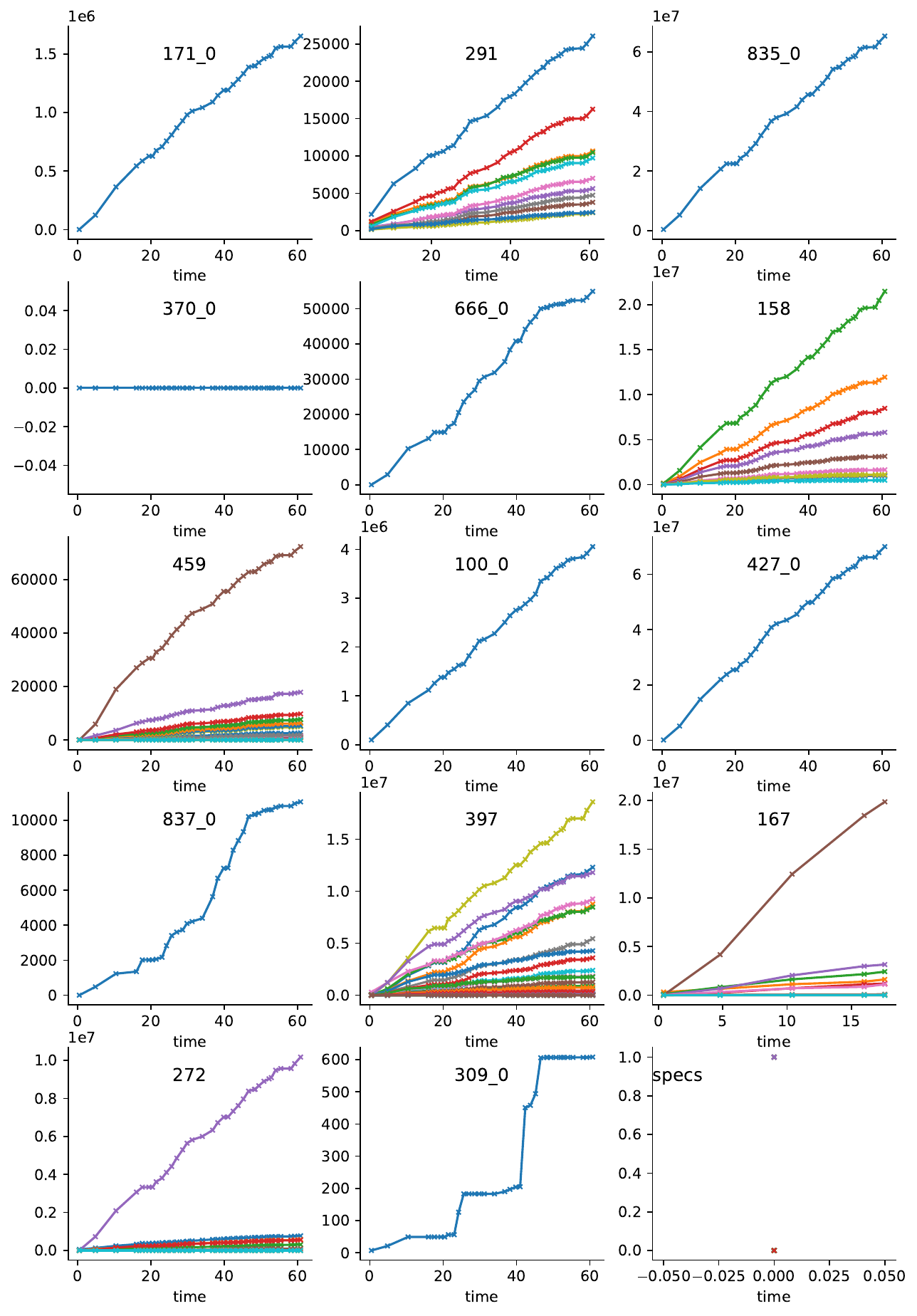}
    \caption{Example input multimodal data for a single vehicle. One plot corresponds to one modality. The sampling instants are highlighted with dots.}
    \label{fig:example_data_compx}
\end{figure}

\begin{table}[ht!]
    \centering
    \begin{tabular}{llrr}
\toprule
 &  & Dimension & \# samples \\
Aspect & Modality &  &  \\
\midrule
\multirow[t]{8}{*}{In-home activity} & Kitchen & 1 & 2196 \\
 & Bedroom & 1 & 2162 \\
 & Hallway & 1 & 2148 \\
 & Bathroom & 1 & 2124 \\
 & Fridge Door & 1 & 2113 \\
 & Front Door & 1 & 2099 \\
 & Lounge & 1 & 1944 \\
 & Back Door & 1 & 1813 \\
\cline{1-4}
\multirow[t]{5}{*}{Physiology} & Body Temperature & 2 & 1580 \\
 & Systolic blood pressure & 2 & 353 \\
 & Diastolic blood pressure & 2 & 353 \\
 & Heart rate & 2 & 353 \\
 & Skin Temperature & 2 & 121 \\
\cline{1-4}
\multirow[t]{2}{*}{Sleep mat} & Sleep HR & 2 & 757 \\
 & Sleep RR & 2 & 757 \\
\cline{1-4}
\multirow[t]{3}{*}{Smart scale} & Body weight & 2 & 275 \\
 & Total body water & 2 & 113 \\
 & O/E - muscle mass & 2 & 113 \\
\cline{1-4}
\bottomrule
\end{tabular}

    \caption{Modality description}
    \label{tab:modality_details_tihm}
\end{table}

\begin{figure}[ht!]
    \centering
    \includegraphics[width=0.5\textwidth]{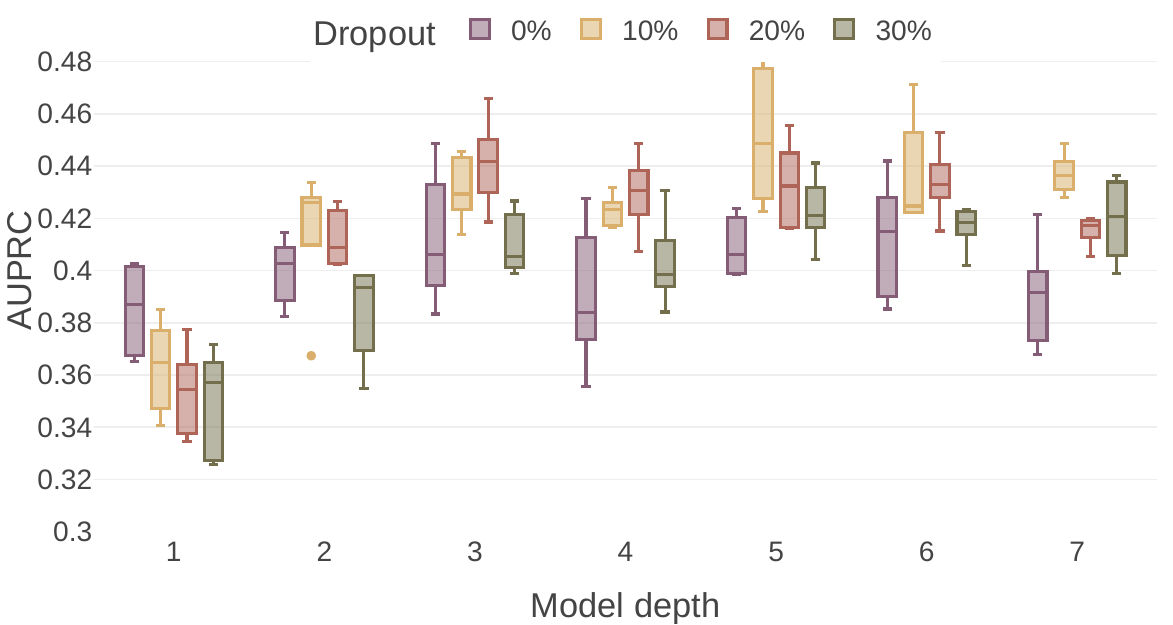}
    \caption{Comparing levels of dropout for various model depth and MLP/1 mixing layers}
    \label{fig:vs_n_layers_mlp1}
\end{figure}

\begin{figure}[ht!]
    \centering
    \includegraphics[width=0.5\textwidth]{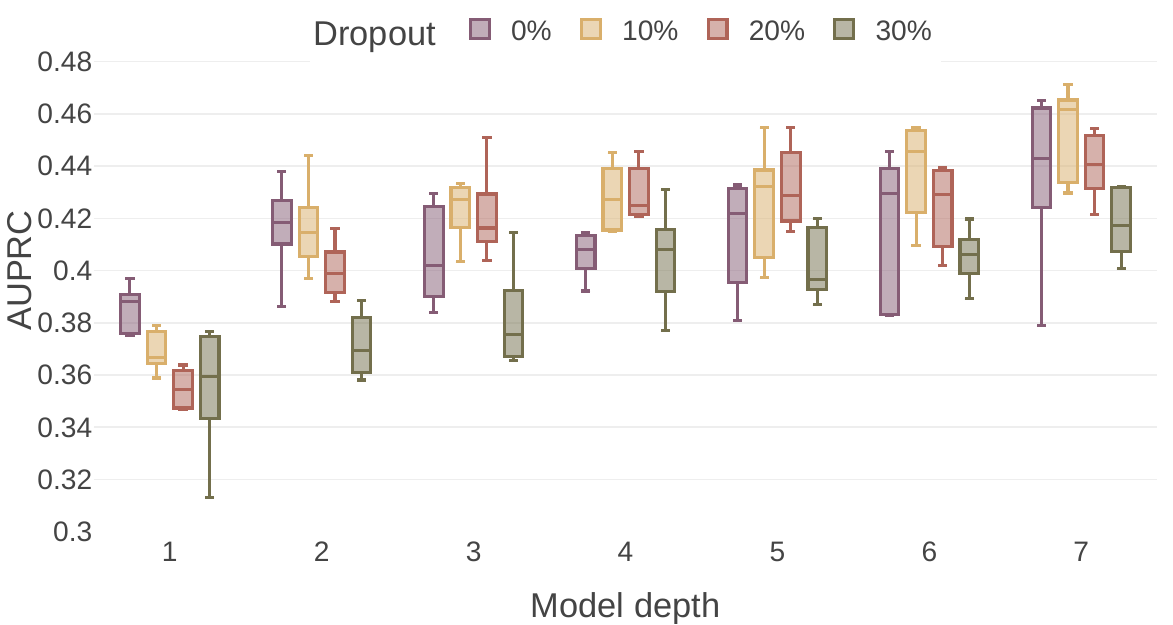}
    \caption{Comparing levels of dropout for various model depth and MLP/2 mixing layers}
    \label{fig:vs_n_layers_mlp2}
\end{figure}

\end{document}